# Diffusion-based Time Series Forecasting for Sewerage Systems

Nicholas A. Pearson*[1], Francesca Cairoli[1], Luca Bortolussi[1], Davide Russo[2] & Francesca Zanello[2]

1 University of Trieste, Department of Mathematics, Informatics and Geosciences, Trieste, Italy
2 Idrostudi srl, Trieste, Italy
*Corresponding author email: nicholasandrea.pearson@phd.units.it

## Abstract

We introduce a novel deep learning approach that harnesses the power of generative artificial intelligence to enhance the accuracy of contextual forecasting in sewerage systems. By developing a diffusion-based model that processes multivariate time series data, our system excels at capturing complex correlations across diverse environmental signals, enabling robust predictions even during extreme weather events. To strengthen the model's reliability, we further calibrate its predictions with a conformal inference technique, tailored for probabilistic time series data, ensuring that the resulting prediction intervals are statistically reliable and cover the true target values with a desired confidence level. Our empirical tests on real sewerage system data confirm the model's exceptional capability to deliver reliable contextual predictions, maintaining accuracy even under severe weather conditions.

## Introduction
The growth of urban centres, both in terms of extension and population size, has brought increased stress on sewerage and drainage systems. These infrastructures are essential for local communities but rely on aging networks which struggle to cope with the effects of increased population in urban areas and widespread soil sealing combined with extreme weather events, with the latter becoming ever more frequent due to climate change (IPCC, 2023). For over 15 years Idrostudi srl, one of Europe's foremost hydraulic engineering professional services consulting firms, has been helping water utilities improve and optimize the management and operation of the sewerage systems they are responsible for, through its expertise in hydraulic and environmental engineering, monitoring, modelling and design. Establishing a monitoring network is essential for understanding the hydraulic functioning of a sewerage/drainage system; however the time series it collects may be inaccurate or unavailable due to anomalies or technical failures (Bertrand-Krajewski et al., 2021). While human experts can identify problematic behaviours upon visual inspection, performing this task on a large scale is often impractical. More complex tasks, such as accurately imputing missing data, pose an even greater challenge without the aid of an automated system. Furthermore, having the ability to perform short to medium term forecasts on network behaviour may be an invaluable support asset for operators on the field, especially during extreme weather events when urban flooding and other undesirable events are more likely to occur. This work focuses on using generative AI to improve the monitoring of the sewerage network: we developed a diffusion model which processes time series data (Tashiro et al., 2021) to deliver probabilistic predictions on the network's behaviour and effectively addresses the aforementioned problems, providing high accuracy and statistically valid coverage guarantees.

## Methodology
Data consists of multivariate time series recording the variations of several physical parameters in specific network sites and precipitation changes over time in the area of interest. The main objective of this work is to perform contextual forecasting, a predictive approach that generates forecasts using historical values of target variables while leveraging both historical and known current information from other variables as additional context. Contextual forecasting can be framed as a time series imputation problem, i.e. the task of estimating the missing values of a time series given its observed ones. This can be achieved by considering the final values of the features of interest as missing values to be imputed, while utilizing the entirety of the others as context. With their work on Conditional Score-based Diffusion Models for Probabilistic Time Series Imputation (CSDI), Tashiro et al. (2021)

proposed a novel approach that leverages Denoising Diffusion Probabilistic Models (DDPMs) (Ho et al., 2020) specifically for the task of time series imputation. DDPMs are a type of deep generative models trained in a two-step procedure. In the initial forward process, existing data is gradually corrupted by iteratively adding small amounts of random noise. In the ensuing backwards process, a neural network learns a denoising procedure that iteratively and gradually removes noise, eventually generating new data starting from pure random noise. At training time CSDI learns an estimate for the conditional probability distribution of the missing values of the time series conditioned on the observed ones. This is achieved using a self-supervised procedure where various portions of the observed values are masked or obscured and the model attempts to reconstruct them given the remaining observations. The choice of which parts to mask plays an essential role in defining what data patterns the model will be able to reconstruct in the inference phase and should ideally match the objective of the target application and the expected missing data patterns. When dealing with contextual forecasting, an appropriate choice could be a random number of final values of the time series for one or more features. This increases model flexibility as, if trained correctly, a single model can impute the targeted data for any combination of concurrently masked features. At inference time the model samples from the estimated conditional probability distribution, transforming random noise samples into plausible imputations. By repeated sampling, we can calculate summary statistics such as the mean or median of the predictive distribution, or determine an empirical quantile range. Given an arbitrary confidence level $\alpha$, the objective is to obtain confidence intervals satisfying the coverage requirement, i.e. containing the true value with a probability greater than $\alpha$. However, intervals constructed from empirical quantiles do not offer this coverage guarantee. We address this limitation using conformal prediction (CP), a flexible approach that adjusts the output of any model, transforming it into predictive intervals with statistically rigorous coverage guarantees. A non-conformity score is computed for each prediction obtained from a held-out calibration dataset by measuring the differences between the predicted values and the ground-truth. The $\alpha^{th}$ critical non-conformity score is then extracted and used as a threshold that, when added to and subtracted from new predictions, provides intervals that entail the desired coverage probability. The choice of which non-conformity score is used, combined with the quality of predictions by the original model, directly influence the size of the intervals, which we aim to minimize to provide meaningful insights. CP requires no retraining of the original model and simply operates as a wrapper around it, only requiring minor adjustments to adapt to the predictive model at hand. Our probabilistic model over time series data calls for CopulaCPTS (Sun and Yu, 2023), a copula-based approach for efficient CP on multi-variate time series. We adapt CopulaCPTS to handle the probabilistic output of CSDI, constructing intervals starting from empirical quantile ranges. In particular, we borrow the non-conformity score from Romano et al. (2019). These intervals provide coverage guarantees, ensuring that the probability that the true values of the time series fall entirely within the prediction intervals is above a desired confidence level. The pipeline of the proposed methodology is fully illustrated in Figure 1.

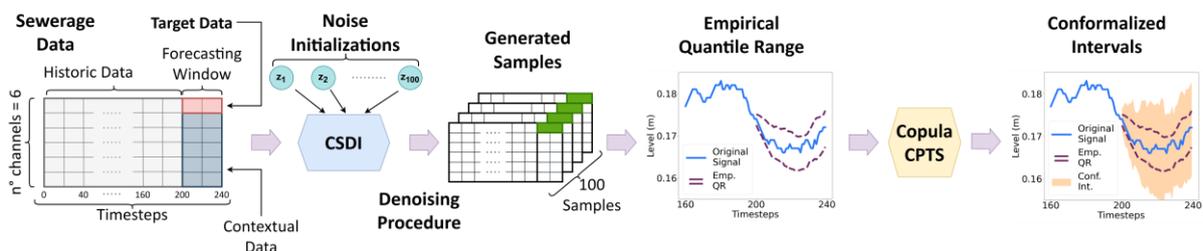

**Figure 1.** Pipeline for the proposed methodology: starting from 100 noise initializations, the CSDI model produces as many plausible imputations, which are used to compute empirical quantile ranges. These ranges are then adjusted using CopulaCPTS to provide intervals guaranteed to cover all true values of the time series with a specified confidence level.

## Case study

We consider a case study focusing on sewerage systems, having access to over four years of validated data from multiple urban sewerage network locations across North-western Italy. These sensors are installed in various pipes and measure the level of sewage every six minutes. In order to leverage spatial correlations between the signals, we select two subsets of five sensors, referred to as Cluster A and Cluster B in the following, which were grouped together based on their topological connections within the same network. The data of each cluster is further enriched with rain intensity measurements taken from rain gauges located in geographical proximity to the acquisition devices of each group. In practical terms, we focus on time series with six channels or features, five for the level measurements taken from the sensors in the cluster and one for the rain intensity, over 240 timesteps, representing a 24-hour period with a recording taken every six minutes.

## Results and discussion

We present experimental results focusing on the imputation of a single channel at the time per cluster, providing estimates for a four-hour time window to emulate the frequency of data transmission from the dataloggers connected to the sensors. Held-out cluster-specific calibration datasets are initially partitioned based on weather patterns to provide condition-specific adjustments to the predicted empirical quantile ranges. This partition is necessary given that the recorded time series generally presents values which are significantly higher during rainfall events compared to dry periods. Combining series with such different scales would have negatively affected the conformal prediction procedure, resulting in intervals which are over-conservative in dry conditions while being insufficiently wide during rainfall events. A single CSDI model is used to generate 100 samples over a test dataset of roughly 600 time series per cluster. Although the same model is used to generate these samples regardless of weather condition or cluster, the test sets are further partitioned using the same criteria as above, resulting in approximately 550 observations per group being classified as dry conditions while 50 series as wet conditions. The test set partition does not impact sample generation but allows for more meaningful result comparisons and for the application of condition-specific adjustments in the CP step. We use the median as an aggregative statistic for the predictive distribution and evaluate its accuracy with respect to the original level value using traditional metrics such as the Mean Absolute Error (MAE) and Mean Absolute Percentage Error (MAPE). Table 1 shows systematic differences between the error rates in dry and wet conditions across both clusters. This is expected as the wet occurrences are less numerous within the dataset and additionally present more heterogeneous dynamics within each event, making them harder to accurately predict. Given a confidence level $\alpha = 0.9$, we retrieved the symmetric empirical quantiles $q_{0.05}$ and $q_{0.95}$ from the set of samples, constructing a quantile range which was further adjusted leveraging the CopulaCPTS procedure. The resulting conformalized intervals provide a 90% guarantee of containing the true values across all timesteps simultaneously. Table 1 also shows that the conformalized intervals provide a significant increase in coverage compared to the empirical quantile ranges for both clusters and weather conditions. These improvements require only minor adjustments to the quantile ranges, with smaller corrections in dry conditions, as shown in the left panel of Figure 2. The increase in coverage can be appreciated in the right panel of Figure 2, where the conformalized interval successfully encompasses the true values of the series over all timesteps, while the empirical quantile range fails to cover the entire signal. Although they still achieve similar coverage levels in all conditions, the conformalized intervals for the sensor of Cluster A appear to be moderately narrower than those for Cluster B. This is particularly noticeable in dry conditions and suggests that the set of samples generated in this scenario is less accurate, as is also visible from the higher error rates in Table 1. The lower accuracy can be attributed to noisier data for the sensor of Cluster B and, as a consequence, the empirical quantiles derived for it require larger adjustments to reach the desired level of confidence. Nevertheless, when evaluated next to the scale of the target values, all of the widths of the intervals in the different scenarios remain within acceptable bounds for practical applications. Lastly, it is important to note that the adjustments performed on the empirical quantiles strictly depend on multiple factors, including the cluster and sensor for which an imputation is being performed and current weather condition. Changing any of these parameters requires using the correct combination-specific adjustment coefficients or, if not already available, repeating the CopulaCPTS process to compute them. Given a set of generated samples for a sufficiently large calibration set, the computational cost of this operation is modest, with the fitting of adjustment coefficients taking only a few seconds, which can then be applied instantly to any new empirical quantile range under the same conditions. Although results have only been presented for two clusters, this methodology has been successfully tested on over ten clusters from the same sewerage network as above, for varying forecasting horizons and on data from other wastewater systems with similar results overall.

**Table 1.** Comparison of various results for two sensors belonging respectively to Cluster A and Cluster B under different weather conditions. The MAE and MAPE error metrics are reported to evaluate the quality of the generated samples (lower values are preferable). Coverage percentages and interval widths are reported to compare empirical quantile ranges against conformalized intervals, along with their per-step average adjustments. Additionally, average target values with standard deviation are shown for each scenario to facilitate result interpretation.

|  | Avg. Target Value (±std) | Error Metrics | | Empirical QR | | Conformalized Intervals | | |
|---|---|---|---|---|---|---|---|---|
|  |  | Avg. MAE | Avg. MAPE | Coverage (%) | Avg. Width | Coverage (%) | Avg. Width | Avg. Correction |
| Cluster A – Sensor 1 – Dry | 0.155 (±0.021) | 0.0026 | 1.6431 | 14.10 | 0.0079 | 88.38 | 0.0179 | 0.0050 |
| Cluster A – Sensor 1 – Wet | 0.284 (±0.160) | 0.0218 | 6.2571 | 27.27 | 0.0718 | 87.88 | 0.2012 | 0.0647 |
| Cluster B – Sensor 5 – Dry | 0.155 (±0.032) | 0.0033 | 2.1275 | 08.07 | 0.0090 | 87.60 | 0.0247 | 0.0079 |
| Cluster B – Sensor 5 – Wet | 0.446 (±0.361) | 0.0305 | 6.2797 | 17.14 | 0.0922 | 88.57 | 0.3259 | 0.1169 |

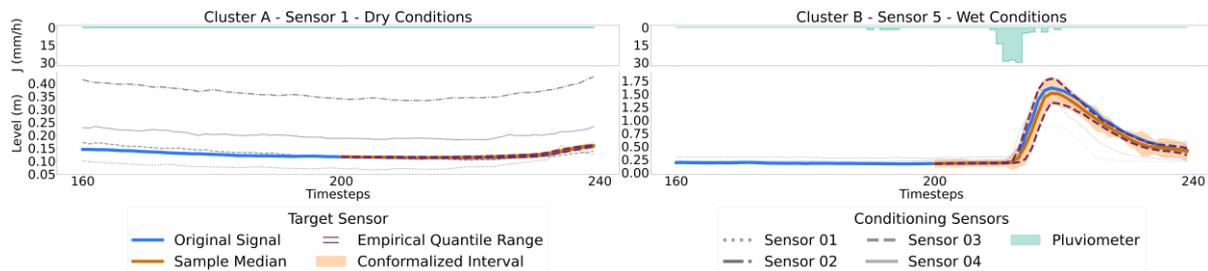

**Figure 2.** Comparison of model output in dry (left) and heavy rain (right) conditions. Both examples show the final 8-hours of a 24-hours period for which a 4-hour imputation is performed. Top panels show precipitation intensity. In dry conditions all samples coalesce to the ground-truth requiring small adjustments. In wet conditions larger corrections are needed.

## Conclusions and future work

This paper presents a novel application of diffusion-based models to perform contextual forecasting in the domain of sewerage systems. Experimental results show that our model performs well both in dry and wet conditions, providing intervals with statistically sound coverage guarantees. This work offers a strong starting point for more complex objectives such as near-real time anomaly detection, pure forecasting and the inpainting of missing data, proving its potential use in sewerage system monitoring through data-driven decision making.